\documentclass[runningheads]{llncs}
\usepackage{graphicx}
\usepackage{amsmath,amssymb} 
\usepackage{color}
\usepackage[noend]{algpseudocode}
\usepackage{algorithmicx,algorithm}

\begin{document}
\pagestyle{headings}
\mainmatter

\def\ACCV20SubNumber{862}  

\title{Online Knowledge Distillation via Multi-branch Diversity Enhancement} 
\titlerunning{Online Knowledge Distillation via Multi-branch Diversity Enhancement}

\author{Zheng Li\inst{1,2} \and
		Ying Huang\inst{1,3} \and
		Defang Chen\inst{4} \and \\
		Tianren Luo\inst{1} \and
		Ning Cai\inst{1} \and
		Zhigeng Pan\inst{1}\thanks{Corresponding author}}
	
\authorrunning{L. Zheng et al}

\institute{
	Virtual Reality and Intelligent Systems Research Institute,\\
	Hangzhou Normal University \and
	School of Information Science \& Engineering, Hangzhou Normal University \and
	Alibaba business school, Hangzhou Normal University\\
	\email{\{lizheng1, caining, luotianren\}@stu.hznu.edu.cn, \\ \{yw52, zgpan\}@hznu.edu.cn},\and
	College of Computer Science, Zhejiang University\\
	\email{defchern@zju.edu.cn} 
}

\maketitle

\begin{abstract}

Knowledge distillation is an effective method to transfer the knowledge from the cumbersome teacher model to the lightweight student model. Online knowledge distillation uses the ensemble prediction results of multiple student models as soft targets to train each student model. However, the homogenization problem will lead to difficulty in further improving model performance. In this work, we propose a new distillation method to enhance the diversity among multiple student models. We introduce {\textbf Feature} {\textbf Fusion} {\textbf Module} ({\bf FFM}), which improves the performance of the attention mechanism in the network by integrating rich semantic information contained in the last block of multiple student models. Furthermore, we use the {\textbf Classifier} {\textbf Diversification}({\bf CD}) loss function to strengthen the differences between the student models and deliver a better ensemble result. Extensive experiments proved that our method significantly enhances the diversity among student models and brings better distillation performance. We evaluate our method on three image classification datasets: CIFAR-10/100 and CINIC-10. The results show that our method achieves state-of-the-art performance on these datasets.

\end{abstract}

\label{key}\section{Introduction}

Knowledge distillation\cite{hinton2015distilling}, as one of the key methods in model compression, the distillation process usually starts by training a high-capacity teacher model. A student model will actively learn the soft label or feature representation[11] generated by teacher model. The purpose of distillation is to train a more compact and accurate student model through the knowledge transferred from the teacher network. In recent years, the convolutional neural network has made very impressive achievements in many vision tasks\cite{krizhevsky2012imagenet,simonyan2014very,he2016deep,long2015fully,redmon2016you}. But it requires high cost of computation and memory in inference, making the deployment of CNN difficult in resource-limited mobile devices. Knowledge distillation was proposed to solve these problems. In the meantime, other types of model compression techniques such as network pruning\cite{han2015deep,lebedev2016fast,molchanov2016pruning} and network quantization\cite{rastegari2016xnor,wu2016quantized,wang2019haq} have also been proposed.

Traditional knowledge distillation\cite{romero2014fitnets,zagoruyko2016paying,yim2017gift} is a two-stage process. We should first train a teacher model, then get a student model by distilling the teacher model. Although this approach can obtain a higher quality student model by aligning the predictions of the teacher model, it is still a complex approach that requires more computational resources. Online knowledge distillation\cite{zhang2018deep} successfully simplifies the training process by reducing the need for pretrained teacher model. Existing online knowledge distillation methods\cite{zhu2018knowledge,anil2018large,chen2020online} learns not only from the ground truth labels but also from the ensemble results of multiple branches. We refer to each branch as a separate student model. This method can improve the performance of models with arbitrary capacity and obtain better generalization ability.

Averaging the predictions of each branch is a very simple way to get the ensemble results. This approach tend to cause branches to quickly homogenize, hurting the distillation performance\cite{kuncheva2003measures,zhou2012ensemble}. However, \cite{zhu2018knowledge,chen2020online} found that the accuracy of the final result improves if different weights were applied to each peer.  In OKDDip\cite{chen2020online}, this paper introduces the concept of two-level distillation method, builds diverse peers by applying a self-attention mechanism\cite{zhang2019self}. Self-attention in OKDDip needs two fully connected layers separately as transformation matrices to obtain importance scores, which increases the complexity of time and space. In ONE\cite{zhu2018knowledge}, the gate module uses features from the second block of its backbone network as input to generate the importance score of the corresponding branch. However, this feature contains little semantic information which leads to limited improvement in image classification tasks.

In this work, we propose a new distillation strategy to enhance the diversity among branches which can significantly improve the effectiveness of online knowledge distillation. By introducing Feature Fusion Module(FFM) to fuse the features of the last layer of multiple branches, we make full use of the diversity of semantic information contained in multiple branches to improve the attention performance\cite{vaswani2017attention}.
Since a large diversity of branches can help ensemble-based online KD methods achieve better results, inspired by \cite{Saito2017AsymmetricTF}, we propose the CD loss to prevent homogeneity between branches by explicitly forcing their features to be learned orthogonally. This loss function serves as a regularization term to prevent group performance degradation caused by homogenization. Unlike other methods in which all branches converge into similar one. By using our method, each branch keeps their uniqueness. Based on \cite{chen2020online}, a two-level knowledge distillation framework is adopted. We build a network with $\textit{m}$ branches, including \textit{m}-1 auxiliary branches and a group leader. The knowledge generated by these diverse peers will be distilled into the group leader, and the remaining peers will be discarded. In order to reduce the consumption of computing resources, we only keep the group leader as the final deployment model.

Our contributions of this work can be summarized as follows:

\begin{itemize}
	\item We propose Feature Fusion Module(FFM) which can better fuse diverse semantic information from multiple branches and improves the performance of the attention mechanism.
	\item We introduce the Classifier Diversification(CD) loss function. As a regularization term, it effectively reduces the homogenization among branches, improves the accuracy of ensemble results and leads to a better student model.
	\item The extensive experiments and analysis verify that our proposed method can effectively enhance branch diversity and train better student models on different image classification datasets: CIFAR-10/100\cite{krizhevsky2009learning} and CINIC-10\cite{darlow2018cinic}.

\end{itemize}

\section{Related Work}

\subsection{Knowledge Distillation}
Knowledge distillation\cite{hinton2015distilling} has been widely used in many scenarios involving deep learning algorithms, such as virtual experiments in VR, autonomous driving and so on. It provides an useful method that allows the complex teacher model to be compressed to a more lightweight student model by aligning the student model with the teacher model. When training the target model, this method takes advantage of the extra supervisory signal provided by the soft output of the teacher model. there are also many works\cite{romero2014fitnets,zagoruyko2016paying,yim2017gift,ba2014deep} made explorations based on this idea. In FitNets\cite{romero2014fitnets}, the student model attempts to mimic the intermediate representation directly from the teacher network. Attention Transfer\cite{zagoruyko2016paying} transfers an attention map of a teacher model into a student and \cite{yun2020regularizing} proposes a similar method using mean weights. In flow-based knowledge distillation\cite{yim2017gift}, the student is encouraged to mimic the teacher's flow matrices, which are derived from the inner product between feature maps in two layers. \cite{lee2018self} saves the computation by using singular value decomposition to compress feature maps.

There are also innovative works exploring alternatives to the usual student-teacher training paradigm. Generative Adversarial Learning\cite{goodfellow2014generative} is proposed to generate realistic-looking images from random noise using neural networks. The ideas in the adversarial network are applied to knowledge distillation\cite{shen2019meal,xu2017training,heo2019knowledge}. In MEAL\cite{shen2019meal}, the generators were employed to synthesize student features and the discriminator was used to discriminate between teacher and student outputs for the same image. In \cite{heo2019knowledge}, this work adopts adversarial method to discover adversarial samples supporting decision boundary. With the supervision of discriminator, student can better mimic the behavior of teacher model. In addtion, many works\cite{park2019relational,peng2019correlation,tarvainen2017mean,xie2020self} have also explored the relationship between the samples. \cite{park2019relational} propose that similar input pairs in the teacher network tends to produce similar activations in the student network. A few recent papers\cite{xie2020self,furlanello2018born,yang2019snapshot} have shown that models of the same architecture can also be distilled. Snapshot distillation\cite{yang2019snapshot} uses the cyclic learning rate policy, in which the last snapshot of each cycle is used as the teacher for all iterations in the next cycle, and the teacher is used to provide supervision signal.

\subsection{Online Knowledge Distillation}

Traditional knowledge distillation methods have two stages that require a pretrained teacher model to provide soft output for distillation. Different from above complex training methods, several works adopts collaboratively training strategy. Simultaneously training a group of student models based on each other's predictions is an effective single-stage distillation method, which can be a good substitute for pretrained teacher models. Some methods\cite{zhang2018deep,anil2018large} solve this problem. The online knowledge distillation was completed through mutual instruction between two peers\cite{zhang2018deep}. However, the lack of a high-capacity teacher model will decrease the distillation efficiency. In \cite{zhu2018knowledge,song2018collaborative}, each student model learns from the average of the predictions generated by a group of students and obtains a better teacher model effect. ONE found that simply averaging the results would reduce the diversity among students, affecting the training of branch-based models. ONE generates the importance score corresponding to each student through the gate module. By assigning different importance score to each branch, a high-capacity teacher model is constructed, which can leverage knowledge from training data more effectively. OKDDip\cite{chen2020online} proposed the concept of two-level distillation. The ensemble results of auxiliary peer networks were distilled into the group leader. The diversified peer network plays a key role in improving distillation performance.

\section{Online Knowledge Distillation via Multi-branch Diversity Enhancement}

\begin{figure}[t]
	\centering
	\includegraphics[width=120mm]{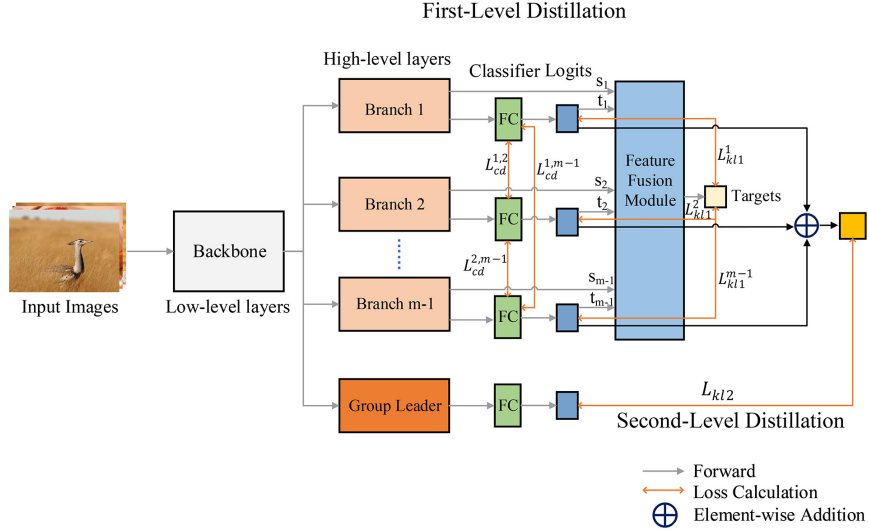} 
	\caption{
		The overall framework of our proposed method. Each branch and shared low-level layers constitute an individual student model. This is a two-level distillation process. For the first-level distillation, each auxiliary branch learns from their ensemble targets. The second-level distillation transfers the knowledge learned by the group to the group leader. $L_{cd}$ denotes the proposed classifier diversification loss. $L_{kl}$ denotes the KL divergence loss. We omit the cross entropy loss $L_{ce}$ for simplicity. We will introduce these loss functions in detail in the third section. Best viewed in color.
	}
	\label{fig:overall}
\end{figure}

The architecture of our proposed method is illustrated in Fig. \ref{fig:overall}. Our method is based on a two-level distillation procedure. The network has $m-1$  auxiliary branches and one group leader. In the first level distillation, each branch learns not only from the ground truth label but also from the weighted ensemble targets obtained through Feature Fusion Module. These results play the role of a teacher model to teach each branch. In the second level distillation, the knowledge learned by the group is further distilled into the group leader. To save computing resources, we use the group leader for the final deployment.

\subsection{Formulation}

In knowledge distillation, the student uses the output of the teacher as an additional supervisory signal for network training. Given a dataset of \textit{N} training samples $D=\lbrace(x_{i},y_{i})\rbrace^{N}_{i}$, where $y_{i}\in\lbrace1,2,...,C\rbrace$. Here, $x_{i}$ is the $i^{th}$ training sample, $y_{i}$ is the corresponding ground truth label and \textit{C} is the total number of image classes. Take the training sample as the input of the teacher network, we will get the output logits $t_{i}=(t^{1}_{i},...,t^{c}_{i})$. The logits vector after softmax will get the $i^{th}$ probability value $q^{j}_{i}$,

\begin{equation}
q^{j}_{i}= \frac{exp(t_{i}/T)}{\sum_{j=1}^C exp(t^{j}_{i}/T)}
\end{equation}
where \textit{T} is the temperature parameter. An increase in the parameter \textit{T} will make the probability distribution smoother. When training teachers, \textit{T} is set to 1. When distilling knowledge from the teacher model to the student model, \textit{T} is usually set to 3.

In order to train a multi-class image classification model, our goal is to minimize the cross entropy between the predicted class probabilities $q_{i}$ and the corresponding ground truth label distribution $y_{i}$,

\begin{equation}
L_{ce}=H(y_{i},q_{i})
\end{equation}
where $H(p,q)=-\sum_{i}p_{i}log q_{i}$.

Knowledge transfer is achieved by aligning the probability distribution $q$ generated by the student with the target distribution $t$. The temperature parameter \textit{T} should be the same for teacher and student networks. Specifically, we use KL(Kullback-Leibler) Divergence as the loss function:

\begin{equation}
L_{kl}=KL(t,q)=\sum_{i,j}t_{ij}log\frac{t_{ij}}{q_{ij}}
\end{equation}

\subsection{Feature Fusion Module}

\begin{figure}[t]
	\centering
	\includegraphics[width=120mm]{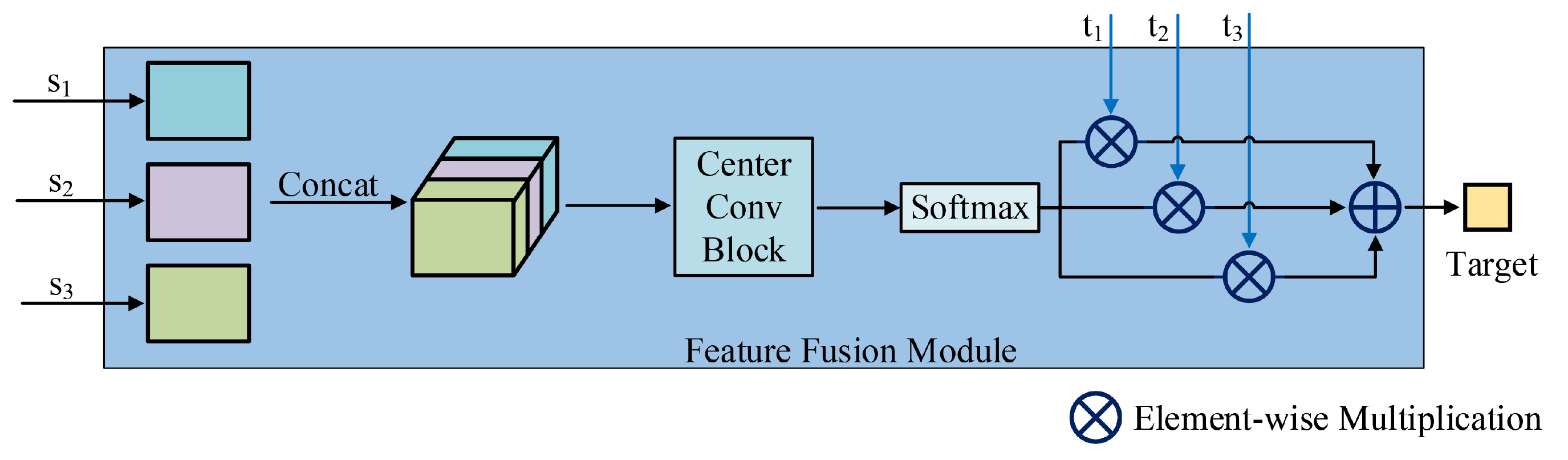} 
	\caption{
		We take the case of three auxiliary branches as an example. Feature map $s_{i}$ from each branch will be concatenated together, and then fed into the center convolution block. The center block is made of several convolutional layers, batch normalization and ReLU activation function. The last layer of this block is the fully connected layer. This block is designed to fuse the semantic representation from multiple branches. Compared with other methods, more semantic information can effectively improve the performance of the module. The final target is obtained by the weighted sum of logits $t_{i}$ of all auxiliary branches.
		}
	\label{fig:ffm2}
\end{figure}

An overview of the Feature Fusion Module is described in Fig.~\ref{fig:ffm2}. Features from a single layer contain less information than features from multiple layers. Many approaches\cite{huang2017densely,ronneberger2015u,li2019dfanet,sun2019deeply} try to take advantage of more diversed features to get better model performance. We take the features of the last block from multiple branches as the input of the Feature Fusion Module. Since deeper layers in the network lead to richer semantic information, this approach can enrich features with high-level semantic information. Our experiment proves that the weights generated from this method can achieve better results.

\begin{equation}
t_{e}=\sum_{i=1}^{m-1}f_{i}(s_{1}, s_{2}, ..., s_{m-1}) \cdot t_{i}
\end{equation}
where $f(\cdot)$ denotes the function of center block in the FFM. This function will output the corresponding importance score for each branch and also satisfy $\sum_{i=1}^{m-1}f_{i}(s_{1}, s_{2}, ..., s_{m-1})=1$. $ s_{i}$ denotes the feature map of the last block from the $i^{th}$ branch. $t_{i}$ denotes the logits from the $i^{th}$ branch. $t_{e}$ denotes the weighted ensemble target.

\subsection{Classifier Diversification Loss}

The diversity has an important influence on the accuracy of the final ensemble results. For better results, we expect peer classifiers to classify samples based on different viewpoints. So we restrict the weight of classifiers, force them to be diversed. We use

\begin{equation}
L_{cd}=\sum_{i=1}^{m-1}\sum_{j=i+1}^{m}L_{cd}^{i,j}=\sum_{i=1}^{m-1}\sum_{j=i+1}^{m}|W_{i}^{T} W_{j}|
\end{equation}
where $W_{i}$ is the fully connected layers' weights of peer classifiers. If the weights of fully connected layers between peers get similar, it means there are more homogenization among them. This loss function acts as a regularization term to prevent homogenization. This will force each classifier to learn different features under this limit. Experiments show that this loss function improves the diversity of peer classifiers and improves the distillation efficiency. We will explain in detail in the ablation study.

\begin{algorithm}[t]
	\caption{Online knowledge distillation via multi-branch diversity enhancement} 
	\hspace*{0.02in} {\bf Input:}
	Training dataset $D$ ; Training Epoch Number $\epsilon$ ; Branch Number $\beta$  \\
	\hspace*{0.02in} {\bf Output:}
	Trained group leader model $\theta^{m}$ and auxiliary models $\{\theta^{i}\}^{m-1}_{i=1}$ \\
	\hspace*{0.02in} {\bf Initialize:}
	e=1; Randomly initialize $\{\theta^{i}\}^{m}_{i=1}$
	
	\begin{algorithmic}[1]
		\While{e $\leq$ $\epsilon$} 
		\State Compute the predictions of all branches $\{\theta^{i}\}^{m}_{i=1}$ with Eq. (1);
		\State Get each auxiliary branch's weight through FFM;
		\State Compute the target logits with Eq. (4);
		\State Compute the CD loss $L_{cd}$ with Eq. (5);
		\State Compute the distillation loss $L_{kl1}$ and $L_{kl2}$ with Eq. (3);
		\State Compute the total loss function with Eq. (8);
		\State Update the model parameters $\{\theta^{i}\}^{m}_{i=1}$
		\State e=e+1
		\EndWhile
		\State \textbf{end while}
	\end{algorithmic}
	
	\hspace*{0.02in} {\bf Model eployment:}
	Use group leader $\theta^{m}$;

\end{algorithm}

\subsection{Loss function and algorithm}

To get a better understanding of our method, we describe the process in Algorithm 1. Our distillation method is a two-level procedure. For the first level distillation, each auxiliary branch learns the knowledge distilled from the soft targets $t_{e}$ generated by FFM. The distillation loss of all auxiliary branches is

\begin{equation}
L_{kl1}=\sum_{i=1}^{m-1}KL(t_{e},q_{i})
\end{equation}

In the second-level distillation, the knowledge learned by the group will be distilled to the group leader. Same as OKDDip, we average the predictions of all branches to get $t_{avg}$. The distillation of the group leader is

\begin{equation}
L_{kl2}=KL(t_{avg},q_{gl})
\end{equation}

To sum up, the loss function of the whole neural network is:

\begin{equation}
L=\sum_{i=1}^{m}L_{ce}^{i}+\alpha T^{2}L_{kl1}+\beta T^{2}L_{kl2}+\gamma L_{cd}
\end{equation}
where $\alpha$, $\beta$ and $\gamma$ are the balance parameter to balance the loss term. The first term is the sum of all branches' cross entropy loss. 

\section{Experiment}

In this section, we evaluate our method on five popular neural networks (ResNet-50, ResNet-110\cite{he2016deep}, ResNext-50(32x4d)\cite{xie2017aggregated}, Xception\cite{chollet2017xception}, ShuffleNet V2-1.0\cite{ma2018shufflenet}) and three image classification benchmark dataset: CIFAR-10/100\cite{krizhevsky2009learning} and CINIC-10\cite{darlow2018cinic}. We also compare our method with closely related works, including ONE and OKDDip. In addition to the classification ability, we also conduct several ablation studies on the feature fusion module and classifier diversification loss, of which the result indicates that the proposed method has better generalization performance compared with other methods. All the reported results are averaged based on three runs.

\subsection{Datasets and Settings}

\textbf{Datasets.} There are three datasets in our experiments. CIFAR-10 and CIFAR-100\cite{krizhevsky2009learning} both contains 50,000 training images and 10,000 test images, which come from 10/100 classes. CINIC-10 consists of images from both CIFAR and ImageNet\cite{deng2009imagenet}. It has 270,000 images and 10 classes. The size in CINIC-10 is the same as in CIFAR. It contains 90,000 training images and 90,000 test images, all at a resolution of 32 x 32. The top-1 classification error rate are reported.

\textbf{Settings.} We implement all the networks and training procedures in Pytorch\cite{paszke2019pytorch}. We conduct all experiments on an NVIDIA GeFore RTX 2080Ti GPU. For all datasets, we follow the experimental setting of \cite{chen2020online}. For data augmentation, we apply standard random crop and horizontal flip to all images. We use SGD\cite{ruder2016overview} as the optimizer with Nesterov momentum 0.9 and weight decay $5e-4$ during training. We set mini-batch size to 128. We use the standard learning schedule. The learning rate starts from 0.1 and divided by 10 at 150 and 225 iterations, for a total of 300 iterations. We set \textit{m}=4, means that there are three auxiliary branches and a group leader. We separate the last two blocks of each backbone network for CIFAR-10/100 and CINIC-10. We empirically set $T$=3 to generate soft predictions. We set $\alpha$=$1$, $\beta$=$2$ and $\gamma$=$5e-8$ to balance the loss term in Equation 6.

We compare our method with several online knowledge distillation methods. In OKDDip, it has two network settings: branch-based and network-based. The branch-based approach refers to student models sharing multiple convolutional layers, separated from each other after a specified layer. The network-based method means that all student models do not share any convolutional layers, and each student is an independent model. The principles of these two approaches are close, so the branch-based method can well validate the effectiveness of our method. In all the experiments, we use branch-based setting for comparison. Baseline means the original model trained on the dataset without any modification. 

\setlength{\tabcolsep}{4pt}
\begin{table}[t]
	\begin{center}
		\caption{
			Error Rate(Top-1, \%) on CIFAR-10.
		}
		\label{table:cifar10_stu}
		\begin{tabular}{l|c|c|c}
			\hline\noalign{\smallskip}
			Models $\quad$& Baseline & Our Method & Gain\\
			\noalign{\smallskip}
			\hline
			ResNet-32    		 & 6.38 $\pm$ 0.10 & \textbf{5.45 $\pm$ 0.07}  & 0.93\\
				
			ResNet-110     		 & 5.46 $\pm$ 0.02 & \textbf{4.47 $\pm$ 0.02}  & 0.99\\	
				
			ResNext-50(32x4d)    & 5.05 $\pm$ 0.12 & \textbf{4.66 $\pm$ 0.05}  & 0.39\\	
				
			Xception		     & 5.70 $\pm$ 0.08 & \textbf{5.19 $\pm$ 0.05}  & 0.51\\	
				
			ShuffleNetV2-1.0     & 9.21 $\pm$ 0.04 & \textbf{8.36 $\pm$ 0.03}  & 0.85\\	
			\hline
			\noalign{\smallskip}
		\end{tabular}
	\end{center}
\end{table}
\setlength{\tabcolsep}{1.5pt}

\setlength{\tabcolsep}{4pt}
\begin{table}[t]
	\begin{center}
		\caption{
			Error Rate(Top-1, \%) on CIFAR-100.
		}
		\label{table:cifar100_stu}
		\begin{tabular}{l|l|l|l|l}
			\hline\noalign{\smallskip}
			Models $\quad$& Baseline $\quad$& ONE $\quad$& OKDDip$\quad$& Our Method \\
			\noalign{\smallskip}
			\hline
			\noalign{\smallskip}
			ResNet-32          & 28.39 $\pm$ 0.04    & 25.76 $\pm$ 0.04	& 25.45 $\pm$ 0.10	& {\bf 24.84 $\pm$ 0.06} \\
			ResNet-110		   & 23.85 $\pm$ 0.17    & 21.94 $\pm$ 0.13	& 21.01 $\pm$ 0.16	& {\bf 20.52 $\pm$ 0.13} \\
			ResNext-50(32x4d)  & 20.43 $\pm$ 0.19 	 & 18.24 $\pm$ 0.03	& 17.90 $\pm$ 0.06	& {\bf 17.55 $\pm$ 0.06} \\
			Xception           & 21.71 $\pm$ 0.06    & 19.69 $\pm$ 0.06	& 19.66 $\pm$ 0.07	& {\bf 19.55 $\pm$ 0.11} \\
			ShuffleNetV2-1.0   & 28.76 $\pm$ 0.12    & 25.23 $\pm$ 0.11	& 25.28 $\pm$ 0.18	& {\bf 25.17 $\pm$ 0.10} \\
			\hline
		\end{tabular}
	\end{center}
\end{table}
\setlength{\tabcolsep}{1.5pt}

\subsection{Results on CIFAR-10/100}

Table~\ref{table:cifar10_stu} and Table~\ref{table:cifar100_stu} compares the top-1 classification error rate on CIFAR-10 and CIFAR-100 based on five different backbone networks. The result generated by ONE is the averaged accuracy of all branches. The results of OKDDip and ours are the accuracy of the group leader. From these two tables, it clearly shows that our method achieves a lower error rate on the same backbone network. Specifically, our method improves the accuracy of various baseline network by 3\% to 4\% on CIFAR-100. The network with higher capacity generally benefits more from our method. Our methods improves the state-of-the-art methods by 0.61\%, 0.49\% and 0.35\% with ResNet-32, ResNet-110 and ResNext-50, respectively. These results showing that our method is more effective than existing methods. When the baseline model has lower capacity, our method can also slightly improve the accuracy compared with other methods.

\setlength{\tabcolsep}{4pt}
\begin{table}
	\begin{center}
		\caption{
			Error Rate(Top-1, \%) of ensemble results on CIFAR-100.
		}
		\label{table:cifar100_ensemble}
		\begin{tabular}{l|c|c|c}
			\hline\noalign{\smallskip}
			Models $\quad$& OKDDip & Our Method & Gain\\
			\noalign{\smallskip}
			\hline
			ResNet-32    		 & 23.22      & \textbf{22.63}  & 0.59\\
			
			ResNet-110     		 & 19.42      & \textbf{18.85}  & 0.57\\	
			
			ResNext-50(32x4d)    & 17.02      & \textbf{16.68}  & 0.34\\	
			\hline
			\noalign{\smallskip}		
		\end{tabular}
	\end{center}
\end{table}
\setlength{\tabcolsep}{1.5pt}

In Table~\ref{table:cifar100_ensemble}, we compare our method with another two-level distillation method OKDDip on three backbone networks. The results of compared methods are the averaged ensemble results of three branches on three backbone networks in the second-level distillation. Since the ensemble results act as a teacher to teach the group leader, a more accurate result can train a better group leader. It is also seen that our method improves the OKDDip method by 0.59\%, 0.57\% and 0.34\% with ResNet-32, ResNet-110 and ResNext-50. Generally, our method successfully enhanced the diversity among different branches and brings improvement to distillation performance.

\textbf{Diversity Measurement.}
We use the interrater agreement in \cite{zhou2012ensemble} as the metric to measure the branch diversity. This method is defined as:
\begin{equation}
s=1-\frac{\frac{1}{T}\sum_{k=1}^{m}\rho({x_{k}})(T-\rho({x_{k}}))}{m(T-1)\bar{p}(1-\bar{p})}
\end{equation}
where $T$ is the total number of classifiers, $\rho({x_k})$ is the number of classifiers that classify $x$ correctly, $\bar{p}$ is the average accuracy of individual classifiers and $m$ is the total number of test samples. OKDDip and our method obtained 0.633 and 0.549 respectively (CIFAR-100 \& ResNet-32). The smaller the $s$ measurement, the larger the diversity. From this results, we can see that our method actually increase the branch diversity.

\subsection{Results on CINIC-10}

CINIC-10 dataset is larger and more challenging than CIFAR-10 but not as difficult as ImageNet. We adopt the same data preprocessing as those of CIFAR-10/100 experiments.

\setlength{\tabcolsep}{4pt}
\begin{table}
	\begin{center}
		\caption{
			Error Rates(Top-1, \%) on CINIC-10.
		}
		\label{table:cinic10_stu}
		\begin{tabular}{l|l|l|l|l}
			\hline\noalign{\smallskip}
			Models $\quad$& Baseline $\quad$& ONE $\quad$& OKDDip$\quad$& Our Method \\
			\noalign{\smallskip}
			\hline
			\noalign{\smallskip}
			ResNet-32          & 15.96 $\pm$ 0.13     & 14.60  $\pm$ 0.09    & 14.41 $\pm$ 0.10	& {\bf 14.28 $\pm$ 0.12 } \\
			ResNet-110		   & 13.99 $\pm$ 0.06     & 12.29  $\pm$ 0.09	 & 12.21 $\pm$ 0.11	& {\bf 11.86 $\pm$ 0.08 } \\
			ResNext-50(32x4d)  & 13.65 $\pm$ 0.12     & 12.19  $\pm$ 0.04    & 12.20 $\pm$ 0.06	& {\bf 12.02 $\pm$ 0.07 } \\
			\hline
		\end{tabular}
	\end{center}
\end{table}
\setlength{\tabcolsep}{1.5pt}

\setlength{\tabcolsep}{4pt}
\begin{table}
	\begin{center}
		\caption{
			Error Rates(Top-1, \%) of ensemble results on CINIC-10.
		}
		\label{table:cinic10_ensemble}
		\begin{tabular}{l|c|c|c}
			\hline\noalign{\smallskip}
			Models $\quad$& OKDDip & Our Method & Gain\\
			\noalign{\smallskip}
			\hline
			ResNet-32    		 & 13.55      & \textbf{13.44}  & 0.11\\
				
			ResNet-110     		 & 11.35      & \textbf{10.98}  & 0.37\\	
				
			ResNext-50(32x4d)    & 11.77      & \textbf{11.54}  & 0.23\\	
			\hline
			\noalign{\smallskip}
		\end{tabular}
	\end{center}
\end{table}
\setlength{\tabcolsep}{1.5pt}

Table~\ref{table:cinic10_stu} compares the top-1 classification error rates based on three backbone networks trained by different methods. From this table, we observed that our method outperforms baseline by 1.68\%, 2.13\% and 1.63\% on ResNet-32, ResNet-110 and ResNext-50 respectively. Our method also improves the state-of-the-art method by 0.13\%, 0.35\% and 0.18\% on three backbone networks. We can find that the improvement in generalization performance is very limited on this dataset. High-capacity networks tend to perform better. But the accuracy of ResNext-50 is slightly lower than ResNet-110 although its baseline performance is better. 

In Table~\ref{table:cinic10_ensemble}, we compare our method with OKDDip. We can find that our method outperforms OKDDip by 0.11\%, 0.37\% and 0.23\% on ResNet-32, ResNet-110 and ResNext-50. While it can be observed that all the methods seem not to increase as much as that in CIFAR-100 experiments. We guess it is because the homogenization problem becomes serious when we conduct experiments on easier datasets. We still need to explore solutions to solve the homogenization problem in the future. 

\subsection{Ablation Study}

\setlength{\tabcolsep}{4pt}
\begin{table}
	\begin{center}
		\caption{
			Ablation Study: Error rates(Top-1, \%) for ResNet-32 on CIFAR-100.
		}
		\label{table:ablation_attention}
		\begin{tabular}{c|c|c|c|c|c}
			\hline\noalign{\smallskip}
			Gate & SA & FFM & CD  & Top-1 error & Top-5 error \\
			\noalign{\smallskip}
			\hline
			\noalign{\smallskip}
			&    	     & \checkmark   &   		  &  25.40      & 6.19   \\
			& \checkmark &			    &   		  &  25.45      & 6.33   \\
			\checkmark	&    	     &    			&   		  &  25.76      & 6.39   \\
			& 		     & \checkmark	&  \checkmark &  \textbf{24.84}  & \textbf{6.08} \\
			& \checkmark &  			&  \checkmark &  25.18      & 6.10   \\
			\checkmark  &    	  	 &  			&  \checkmark &  25.31      & 6.11   \\
			\hline
			\noalign{\smallskip}
		\end{tabular}
	\end{center}
\end{table}
\setlength{\tabcolsep}{1.5pt}

In this section, we conduct various ablation studies to validate the effectiveness of our proposed FFM and CD loss. We use ResNet-32 on the CIFAR-100 dataset to show the benefit of our components. We also compare our FFM with other knowledge distillation methods, including gate module in ONE and self-attention(SA) mechanism in OKDDip. 

In Table~\ref{table:ablation_attention}, we report the top-1 and top-5 error rates of different methods. The remaining experimental settings are consistent with previous experiment. We carefully conducted six experiments on the network components. We compared the performance of three attention modules in the same experimental settings. When FFM is used only, the performance of our method has slightly exceeded other methods. This shows that FFM makes the student network learns more knowledge during the distillation. Compared with gate module in ONE, our method improves the top-1 error rates by 0.36\% and top-5 error rates by 0.2\%. This result proves that our method effectively utilizes the rich semantic information of multiple branches. When we combine different attention mechanism with classifier diversification loss, our results clearly show that our method surpasses other methods. The combination of FFM and CD loss has more obvious improvement. Compared with the independent FFM, the combination improves the top-1 error rates by 0.56\% and the top-5 error rates by 0.08\%. Our method clearly enhances the diversity among branches and improves the generalization ability of the student model. From this table, we observe that CD loss really plays the most important role in the overall improvements.

\begin{figure}[t]
	\centering
	\includegraphics[width=120mm]{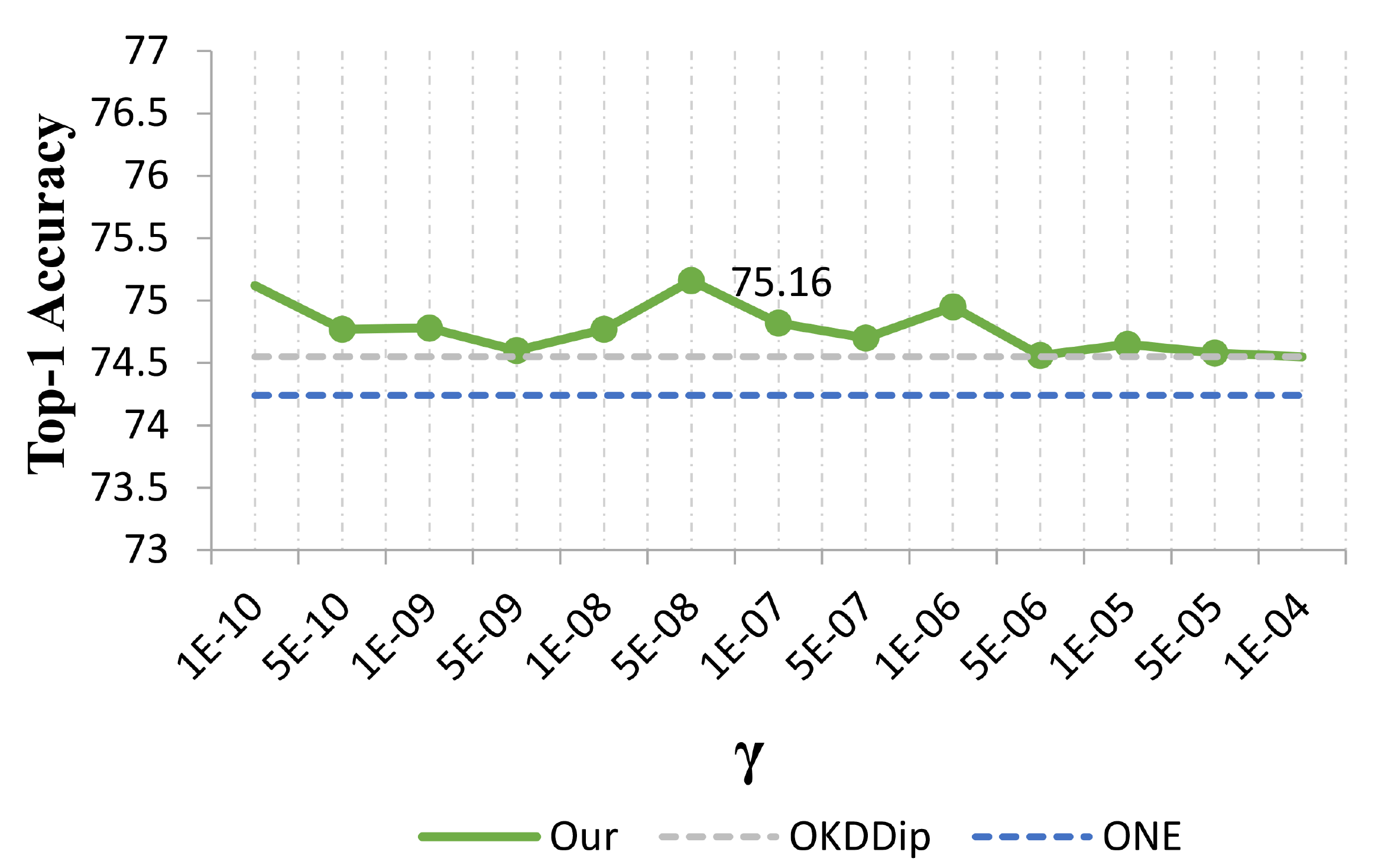} 
	\caption{
		Sensitivity to $\gamma$ on CIFAR-100 for ResNet-32.
	}
	\label{fig:ablationstudy_gamma}
\end{figure}

Fig. \ref{fig:ablationstudy_gamma} demonstrates how the performance of our method is affected by the choice of hyperparameter $\gamma$ of the CD loss. We plot the top-1 accuracy on the CIFAR-100 for ResNet-32 group leader trained with $\gamma$ ranging from $1e-10$ to $1e-4$. In this figure, the dash line indicates the mean accuracy of other methods. We can find that our method still has robust performance against varying $\gamma$ values. The green dot indicates the parameter we are using. We should note that the choice of parameters will affect the optimization process. If the parameter is too large, this will lead to too much diversity among the branches, and eventually will not converge. If the parameter is too small, the CD loss function will be difficult to play the role of regularization. In that case, the value of this loss function will be very small, making the loss function ineffective. This figure shows that CD loss has a significant effect on distillation performance within a proper range.

\section{Conclusion}

In online knowledge distillation, diversity is always an important and challenging issue. In this work, we proposed the Feature Fusion Module and Classifier Diversification loss, which effectively enhances the diversity among multiple branches. By increasing branch diversity and using more diversed semantic information, we have significantly improved the performance of online knowledge distillation. Experiments show that our method achieves the state-of-the-art performance among several popular datasets without additional training and inference costs.

\section*{Acknowledgement}

This work is supported by National Key Research and Development Project of China (Grant No. 2018YFB1004901).

\bibliographystyle{splncs}
\bibliography{egbib}

\end{document}